# Transcending the "Male Code":
# Implicit Masculine Biases in NLP Contexts

**KATIE SEABORN**

*Tokyo Institute of Technology*
*Tokyo, Japan*

**SHRUTI CHANDRA**

*University of Waterloo*
*Waterloo, Canada*

**THIBAULT FABRE**

*The University of Tokyo*
*Tokyo, Japan*

**ABSTRACT:** Critical scholarship has elevated the problem of gender bias in data sets used to train virtual assistants (VAs). Most work has focused on explicit biases in language, especially against women, girls, femme-identifying people, and genderqueer folk; implicit associations through word embeddings; and limited models of gender and masculinities, especially toxic masculinities, conflation of sex and gender, and a sex/gender binary framing of the masculine as diametric to the feminine. Yet, we must also interrogate how masculinities are "coded" into language and the assumption of "male" as the linguistic default: implicit masculine biases. To this end, we examined two natural language processing (NLP) data sets. We found that when gendered language was present, so were gender biases and especially masculine biases. Moreover, these biases related in nuanced ways to the NLP context. We offer a new dictionary called AVA that covers ambiguous associations between gendered language and the language of VAs.

## 1  Introduction

Machines can now speak with us. Advances in machine learning (ML) and natural language processing (NLP) have paved the way for more natural modes of communication with computers. Virtual assistants (VAs), conversational user interfaces (CUIs), and voice user interfaces (VUIs) in the form of smart speakers embedded in homes, spaces, and cities, can be found alongside social robots that greet us at the storefront. Amazon Alexa, Apple's Siri, Microsoft's Cortana, and LINE's Clova are commercial examples being taken up at a global scale. While speaking with and listening to computers has a long history within the field of human-computer interaction (HCI), these recent industrial and commercial trends have reinvigorated interest in the voice of the machine [28,95]. In SIGCHI and adjacent spaces, this can be seen in recent workshops (e.g., [15,79,99,104]) and conferences (notably CUI[1] and ACL[2]).

Voice, speech, command-based, and conversational interactions are fundamentally about language. Most systems rely on the NLP of dictionaries and dialogue models trained on human (or human-generated) data sets. Many have harnessed the power of the crowd, notably Amazon Mechanical Turk (AMT), to create these data sets. The merits and demerits of people-powered data sets have been well-documented [21,23,26,47,64]. On the one hand, NLP-based systems trained on human language texts can lead to more natural, fluent language capabilities in machines. On the other, they can include biased and harmful content, on purpose or not [1,17,23,27,35,47,102]. Critical work has highlighted explicit and implicit forms of bias in algorithms and data sets for gender [17,23,27,101,102], race [16,77,81,92], age [35], and their intersections [49,77]. Most work has focused on gender stereotypes, harassment, and abusive language, and otherwise limited and/or negative associations with femininities and genderqueer people, notably girls, women, femme, trans femme, and non-binary femme people [23,101,102]. Critical scholarship has also escalated a need to distinguish sex from gender [14,20,34,71] and consider genders and sexes beyond the binary [34,63,98]; refer also to emerging work[3]. Since language is a medium of power [40], this area is a necessary social good; its impact cannot be understated.

---

[1] https://www.conversationaluserinterfaces.org
[2] https://aclanthology.org/venues/acl
[3] https://arxiv.org/abs/2202.11923

Nevertheless, there appears to be a gap when it comes to how the "gender problem" is framed. Most work can be organized under three broad categories: (i) detecting and rectifying stereotypes as limited and/or negative representations of gender within data sets used to train ML algorithms, where the focus has tended to be on women and femininities [23,102]; (ii) tracing the spread of toxicity, harassment, and abusive language, especially related to misogyny, within digital "manospheres" [42,85]; and (iii) debiasing ML algorithms and data sets using methods that tend to operationalize gender bias as the relative distance between terms associated with masculinity and its assumed diametric, femininity, i.e., word embeddings [17,23,102]. Underlying much of this work is a sex/gender binary model of man/woman, male/female, and/or masculine/feminine [34,62]. For example, the famous word embedding of "man is to computer programmer as woman is to homemaker" [17] demonstrates how "masculinity" is implicitly singular and always positioned against its "opposite," femininity, on a bipolar scale [30]. Yet, addressing gender bias also means taking on a broader view of gender, especially of masculinities as a covertly dominant gender [66], as well as covering a broader range of subtle associations that arise in language use [2,45]. Tackling discrimination, toxicity, and stereotypes is crucial, but so is scrutinizing the more subtle varieties of gender bias that crop up in word choices and gendered assumptions.

As a first step, we targeted an underexplored framing of gender bias for NLP-based VA development: whether and to what extent implicit forms of masculine-centric language exist, i.e., implicit masculine biases. Specifically, we aimed at societal tendencies to centre masculinity, consciously or otherwise, in gendered language, i.e., word choices attributed to specific genders, and masculine-as-norm language, i.e., the overrepresentation of masculine references. We asked: Do implicit masculine biases in terms of gendered language and masculine-as-norm language, specifically pronoun use, gender-marked words, and names, exist in the NLP context of data sets created to train VAs and to what extent? To answer this question, we examined two large-scale and freely available VA-oriented data sets—MASSIVE[4] and ReDial[5]—using established dictionaries with automated content analysis and manual thematic analysis. Our contributions are:

(i) A new framing of gender bias for NLP that targets implicit masculine biases in novel ways, including masculine language, masculine-as-norm language, and masculine-by-default patterns,

---

[4] https://github.com/alexa/massive
[5] https://redialdata.github.io/website

(ii) Evidence of gendered language more generally but especially implicit masculine biases in two of the most rigorously developed open NLP data sets available, and

(iii) AVA, a new dictionary that reveals the nuances in and ambiguity of gendered language within VA contexts.

The significance of this work is in framing gender bias in VA-oriented NLP data sets differently, if a bit ironically: centring masculinities and masculine language to interrogate masculine-centrism through two novel lenses on implicit bias, namely gendered language and masculine-as-norm language. We also offer a concerted, if limited, effort to disambiguate sex and gender and move beyond the sex/gender binary. A broader view of masculine biases and gender can reveal overrepresentation and inequities in these data sets. As we found, it also presents opportunities for creating new material on bias detection and debiasing in VA and NLP contexts, i.e., AVA. This view is a necessary extension to the work on gender bias and debiasing linguistic data sets for computers that can understand and speak with us.

## 2   Related Work

### 2.1   GENDER BIASES AND DEBIASING IN NLP

Human language is biased, and language-based data sets are no exception. A growing and critical body of work in NLP has started to recognize and take action on gender and other forms of bias [23,27,102]. Initiatives include: locating and analyzing the extent of biases in NLP data sets [22,31,49]; developing methods by which to debias these data sets and the algorithms that use them [1,17,106]; and crafting alternatives, which can mean bringing in marginalized genders and disrupting ideas about gender norms [101]. Yet, there are notable gaps and even biases in this body of work. Cao and Daumé [20] raised awareness of the distinction between social and linguistic genders, warning those of us working in NLP, if not especially on English data sets, to avoid drawing conclusions of subject gender based on gendered terms. Others have noted a lack of intersectionality [19,49] and gaps in representation of genders beyond the binary [1,102]. We wish to add *masculinities* to this list. We are in no way criticizing these trajectories, which are essential for achieving the elimination of inequalities in NLP and HCI. In fact, we would argue that gender bias in NLP is becoming a core area of feminist HCI [4,5,9,25], given that VAs continue to be taken up as a focus of study and by end-users [28,95]. Nevertheless, fighting sexism and toxicity will require considering gender and language in a different way [84].

When it comes to developing NLP data sets, there is a tension between the goals of realism in dialogues and avoiding gender bias. Many rely on curation and crowdsourcing, especially Amazon Mechanical Turk (AMT) to gather, annotate, and judge data sets. This can result in natural and large data sets, but at the risk of perpetrating gender biases [61]. For example, people tend to assume, when no gender information is available, that undescribed people are men, from characters in books to voters to celebrities [2]. Moreover, women, when primed to think about gender, may be more likely than men to refer to gender in subsequent written work [91]. Experts in NLP have urged caution and advocated for a nuanced approach to evaluating bias and inequities, one that involves recognizing who the *creators* of the data were *and* who *curated* the data, as well as what or who the data is *about* [12,78]. At the least, we can map out the extent of these biases in data sets and then make a conscious decision to act or not, and how. This is crucial for open data sets that are taken up widescale and used to train the voices of everyday machines.

The notion of "scale" is also shifting. Those working in NLP and adjacent spaces are recognizing how WEIRD (sampled from Western, educated, industrial, rich, democratic nations) most data sets are [23,57,74,111]. NLP data sets are often in English and oriented to the US context. Efforts on translation and localization of NLP data sets have emerged, such as with the MASSIVE data set. As Bardzell outlined years ago for feminist HCI praxis [4], we must consider how these data sets and their biases exist in and influence a larger *ecology* of systems and interactions. Yet, it is not clear whether and how crowdsourced efforts on NLP data sets take gender bias into account. For instance, the preprint about the MASSIVE data set does not discuss bias at all[6]. Without recognition, the danger of (re)encoding biases or introducing new ones is great.

At present, the bulk of the work has framed gender bias as stereotyped gender associations embedded in core NLP features, especially word embeddings based on gendered occupations and social roles [17,49,102], and toxic masculinity, characterized by hate speech, aggression, and misogynistic conduct [42,85,102]. Field and Tsvetkov [45] offer one of the few NLP efforts to tackle a more subtle form: *implicit bias* [52] as predicted by unsupervised ML based on comment addressee, i.e., how likely the addressee is to be a woman. Another issue is the focus on women being stereotyped and the ideal solution being to "flip the script," i.e., using feminine referents where there would be masculine ones. Indeed, "gender-swapping" is a common approach to debiasing [102]. While some have recognized that this excludes non-binary genders [102], it also prescribes a singular view of masculinity, positions masculinity as diametric to femininity, and erases the notion of gender fluidity [20] and multiplicity [63]. Finally, when it comes to toxicity, changing content must be viewed as a sensitive process. Removing expressions of hate may lead to an agent that is not trained to understand and respond in prosocial ways, such as to enlighten naïve interlocutors, like children, or "fight back" against abuse [13,44,100]. Toxic language is also contextual [1]. For example, queer communities may "take back" and use toxic language that is parsed by well-meaning but heteronormative algorithms as toxic [51]. Training VAs involves not only giving these agents a voice, but also allowing them to understand us. As recent work on bias in speech recognition has revealed [68,103], VAs must be trained on diverse data sets to maximize inclusion.

---

[6] https://arxiv.org/abs/2204.08582

## 2.2 OPERATIONALIZING MASCULINE BIASES

We provide an overview of how masculinity has been framed within NLP and HCI work. We then introduce the less-trodden framing in this body of work: implicit masculine biases in language use.

### 2.2.1 *Toxic Masculinity, Misogyny, Sexism, and Stereotypes.*

Much of the work on gender bias has focused on negative forms of masculinities. *Toxic masculinity* means adherence to traditional and limited masculine gender roles whereby men withhold emotions and avoid behaviours deemed *not* masculine and express themselves in ways regarded as masculine, especially anger, aggression, and dominance [55]. In NLP-related work, this has been explored via *toxicity* [42,85]. While the term "toxic masculinity" may not be used (and anyone can partake in toxic behaviours), this body of work is coded masculine through explorations of "the manosphere," or the virtual spaces in which primarily men promote narrow views of masculinity, if not sexism and anti-feminist beliefs. *Misogyny* is a related concept, referring to hatred towards women and/or femininities [76]. It is an extreme form of blatant sexism towards a specific gender. *Sexism* is a broader concept, including stereotypes, benevolent varieties [50], e.g., women have high emotional IQs, and sex/gender-based discrimination against genderqueer people and those outside of the gender binary, e.g., fear of trans folk. Toxicity and sexism are often bundled together in NLP work, though some recognize that not all forms of toxicity are gendered [90]. The majority of other work on gender bias in NLP has focused on *stereotypes*, defined as widely held but limited and often socially harmful ideas about a group of people identified by a gender label [17,27,31,37,102], e.g., women being homemakers. As discussed, "gender-swapping" [102] and "gender evasive" approaches [94] are widely used "debiasing" strategies, but are limited by a gender binary framing. In short, NLP and adjacent domains have approached language-based representations of masculinities in the extremes: toxic manifestations that need to be culled or counterparts to devaluations of femininity, all with a gender-constricted framework. Yet, there are more subtle forms of masculine gender biases at play in the wider world of words.

### 2.2.2 *Implicit Masculine Biases.*

Societies around the world at various times in history have tended to centre men and masculinity, a phenomenon that has been called *androcentrism*, *male-centredness*, *male as norm*, or *male by default* [2,8,11]. In this paper, we use *implicit masculine bias* as an umbrella term to centre our framing as a matter of gender and avoid the conflation of sex and gender in terms such as *andro* and *male*. We therefore also use the terms *masculine-centric language*, *masculine as norm*, and *masculine defaults or masculine by default* [24], as well. This phenomenon of implicit masculine bias is a reflection of how social power operates through gender as a social construct, whereby men and masculinity are positioned at the top of the social hierarchy; this is commonly referred to as a *patriarchal* system [46]. Masculinity and by association men are defined as the power-holders and society is organized around this stance [97]. The repercussions are vast, affecting most aspects of most societies in obvious and subtle ways. For instance, a word embeddings study [3] using a corpus of over 630 billion words from sources on the Internet showed that the "gender neutral" words "person" and "people" are more frequently associated with "man" and "men." We may not use masculine words explicitly, but we mark humanity as masculine and masculine-centric by other means, including when we codify our worldviews in language.

The ways in which implicit masculine biases play out in language vary. One way is the use of unambiguous masculine references as neutral or universal, in presence and extent. As de Beauvoir articulated almost 75 years ago [8], masculine pronouns, possessives, honorifics, and gender-marked words in romantic languages like French and English are used by default and treated as neutral, universal, and "unmarked," meaning that, for most people, they carry no extra meaning in relation to gender. These words are explicitly gendered but also implicitly treated as gender-neutral, thereby positioning masculinity as fundamental and the ideal [2]. For example, when presented with neutral stimuli, people are more likely to assign a masculine gender or use a masculine reference [54]. In contrast, feminine references are used to position "woman" as "other," a special case that is distinct and oppositional to "man." To give an encompassing example: "Guys, he throws like a girl." Notably, masculinity is not typically positioned next to genders beyond the binary or framed in a gender fluid way. Also, gender-marked words and linguistic expressions may be referential, i.e., refer to specific entities, such as "the cowgirl," or lexical, i.e., refer in a more general way to entities that fall under that gender classification, such as "cowgirls are ..." [20]. *Frequency* of masculine references is key, including gendered names, gender-marked words, like "businessman," and masculine pronouns [2]. We seek to answer recent calls in NLP [37] by examining such *masculine-as-norm* or *masculine-by-default language* as one linguistic variety of implicit masculine bias.

Another is *gendered language*, an implicit form of language use that is virtually unexplored in NLP. Gendered language is the idea that *word choice itself* is gendered, where the words people use in speech or to refer to themselves in writing tend to be gendered in ways that reflect their socialization and identity [48,87]. Men, for instance, tend to use dominant language to express masculinity in terms of authority and power [67]. Other identities can intersect with and influence gendered word choice. Roberts and Utych [87] found that Republican presidents of the US tend to use more masculine language than their "fellow" Democrats. Gendered language can be unintentionally and unknowingly deployed, too. Gaucher, Friesen, and Kay [48] found evidence of widescale gendered word choices in job advertisements that resulted in stereotyped effects: masculine language drew in more men and pushed more women away, despite these women knowing that they were qualified. In the only NLP example we could find, Petreski and Hasim [83] discovered masculine-centric biases in Word2Vec word embeddings via the Small World of Words (SWON-EN2018: Preprocessed) dictionary from De Deyne et al. [33]. Yet, the authors took a binary approach that conflated sex and gender and excluded diverse sex and gender participants and language; this work also did not relate to VAs. We attempt to translate this previous research to the context of NLP data sets built for VAs, exploring these notions of the *genderedness of language,* especially *word choice as masculine language*, in a more sensitive, gender-expansive way.

In HCI, we are creating VAs and CAs and CUIs and VUIs and video game characters and social robots that speak with the voices of these data sets. We know that agents can reflect the implicit biases of their creators. Even so, these biases can be reinforced or disrupted on awareness and the choice to act. Given recent efforts, it is possible that new data sets, like MASSIVE, may be less biased than older data sets, like ReDial. If not, we can act now before the models trained on these data sets become widespread and the current versions of the data sets become standard.

## 3   Data Sets

### 3.1   NLP DATA SETS

Most NLP data sets are composed of labelled data. For example, a datum could be a photo of a cat, which may have one or more labels, including "cat," "furry," "tortoiseshell," "adorable," "feline," and so on. Such sets of labelled data are used to train machine learning models. The goal of creating these models is to teach the machine how to recognize new data that are similar to but not quite the same as the data used during training. To evaluate these trained models against data never seen in training, a single data set is usually separated into three parts: "train," the training data; "dev," the data used after one training cycle (of which there are typically many) to check the progress of the model currently being trained; and "test," a selection of data not used in training, i.e., data that the machine has yet to encounter and is "naïve" about, that is used to evaluate the final model. Test data usually contains edge cases to better evaluate the model's performance. MASSIVE uses all three splits, while ReDial provides train and test splits only.

### 3.2   THE MASSIVE DATA SET

We used the MASSIVE data set developed for intelligent voice assistants, notably Amazon Alexa. MASSIVE was released on April 20[th], 2022 and contains over one million utterances in 51 languages across a range of scenarios and intents. The English data set contains a total of 113,979 words. There are 6152 unique non-stemmed[7] words, including stop words, e.g., "and," "the," "on," and pronouns. With stemming[8] and stop words removed, there are 4940 unique tokens. The data set is provided in JSON format with train, dev, and test partitions for ML; for example:

---

[7] Stemming is an inexact process of reducing an instance of a word to its stem, e.g., "running" to "run." Lemmatization or lemming is the process of reducing a word to its root form exactly, i.e., not just removing the "ing" to find the stem but transforming it into its dictionary form, called the lemma.
[8] In quanteda, stemming is done with Snowball: http://snowball.tartarus.org

*{"id": "13371", "locale": "en-US", "partition": "train", "scenario": "qa", "intent": "qa_maths", "utt": "how do you subtract numbers", "annot_utt": "how do you subtract numbers", "worker_id": "269"}*

*{"id": "13373", "locale": "en-US", "partition": "dev", "scenario": "general", "intent": "general_quirky", "utt": "who are the top five all time n. h. l. goal scorers", "annot_utt": "who are the top five all time n. h. l. goal scorers", "worker_id": "263"}*

*{"id": "13372", "locale": "en-US", "partition": "test", "scenario": "general", "intent": "general_quirky", "utt": "what time are the hockey games tonight", "annot_utt": "what time are the hockey games [timeofday : tonight]", "worker_id": "263"}*

MASSIVE is a text-based version of the SLURP data set[9], an English-only voice data set released in 2020 [7]. SLURP, or Spoken Language Understanding Resource Package, was created to diversify audio-to-text data sets and reduce noise-based errors between audio and semantic features. It was created in two phases for the context of an in-home personal robot that can handle many features of daily life. First, commands for the robot were crowdsourced over AMT. Then, over 100 participants were asked to read out these commands in home and office settings for about 1 hour each. SLURP contains over 72k English audio recordings. MASSIVE aimed to build on this data set by localizing SLURP into 50 other languages. This was done using AMT; the full details are provided in a preprint report[10].

We chose MASSIVE for several reasons. First, it is extremely new, representing the latest efforts and quality standards in designing for NLP systems. Second, it was built on large-scale, crowdsourced natural dialogue that was then validated within actual use contexts, lending it a certain level of ecological validity. Third, it was made freely available by a powerful entity and used in a global competition[11]. As it is likely to spread quickly and be used widely, the impact could be widescale. Fourth, it is available in a large range of languages. While the opportunity for global impact is high, this also allows for future analyses and comparisons across linguistic and cultural settings. This may be especially important for matters of gender bias and sexism, especially more subtle varieties, as indicated by cross-cultural work [82]. Fifth, the creators of the data set appear to be open to, if not eager for, evaluations and critical feedback on the data set. We offer a new perspective on evaluating the quality and fairness of the data set in terms of gender.

---

[9] https://github.com/pswietojanski/slurp
[10] https://arxiv.org/abs/2204.08582
[11] The Massively Multilingual NLU 2022 workshop (MMNLU-22 at EMNLP 2022), which started on July 25th, 2022: https://mmnlu-22.github.io

## 3.3   THE REDIAL DATA SET

The ReDial dataset [73] was created for training agents with goal-directed and conversational dialogue in English. ReDial was a multi-party effort, involving Google, IBM, and the National Sciences and Engineering Council of Canada (NSERC), and support from Microsoft Research. The data set is made up of dialogues between two people recommending and talking about movies, especially their thoughts on, feelings about, and desires to find movies. Published in 2018, it contains 11,348 dialogues with 1,371,903 words and 16,261 unique non-stemmed words, including stop words, e.g., "and," "the," "on." With stemming and stop words removed, there are 12,281 unique tokens. The dialogues were crowdsourced by pairs on AMT with an unknown number of people located in Canada, the US, the UK, Australia, and New Zealand. The data set is freely available in JSON format under the CC BY 4.0 License. More details are available in Li et al. [73]. Here is an example from the training data set, formatted and truncated for readability:

> *{"timeOffset": 48, "text": "I saw @119144 last night and really liked it!", "senderWorkerId": 30, "messageId": 3183},*

> *{"timeOffset": 94, "text": "That was a good movie.  If you like Superhero Movies you should check out @169419", "senderWorkerId": 44, "messageId": 3184},*

> *{"timeOffset": 112, "text": "Is that out already? I really wanted to see that one.", "senderWorkerId": 30, "messageId": 3185},*

Like MASSIVE, ReDial offers a large, English-language data set of utterances crowdsourced on AMT and geared towards training VAs. However, it is a relatively older data set (2018 versus 2020/22) released just as momentum on gender bias and debiasing was beginning to take root. As such, it may offer a time-sensitive comparison.

## 4   Study 1: Implicit Gendered Language

We carried out Study 1 in three phases: the initial study (Study 1.1), which led to a dictionary development phase (Interlude), whereafter we reconducted the initial study (Study 1.2) using this new dictionary: AVA. The original procedure was registered on OSF[12] in advance of data analysis on August 7th, 2022 and updated on January 25th, 2023.

---

[12] https://osf.io/dwf5v

## 4.1 METHODS

We analyzed the relative presence of implicit gendered language using automated content analysis, i.e., content analysis performed by computers using NLP methods to extract meaningful patterns from text materials. We used two dictionaries applied to the English (US) subset of the MASSIVE data set and the entire ReDial data set. We supplemented the automated results with a manual contextual analysis to dis/confirm the genderedness of each term.

### 4.1.1 Dictionaries.

An overview of the two dictionaries is in Table 1. These were the only dictionaries on implicit gendered language that we could find. We chose each based on its relevance (does it assess masculine language?), quality (was it developed in a rigorous way?), and peer acceptance (do academics and/or experts in NLP use it?). The dictionaries are diverse: varied in context, date of origin, method of creation, format, number of tokens. However, they are limited to the gender binary, offering feminine and masculine gendering only. We describe each in detail below:

- The dictionary by Roberts and Utych [87] was created to explore the relationship between implicit gendered language and partisanship, developed through crowdsourcing on AMT. The authors asked 175 mturkers to rate the genderedness of words selected from the Affective Norms for English Language (ANEW) data set by Bradley and Lang [18] using a bipolar 7-point scale with a neutral centre. They defined two cut-offs for masculine—greater than or equal to 5 (n=77) or greater than 5.5 (n=30)—and feminine—less than or equal to 3 (n=36) or less than 2.5 (n=8)—with 4 as the gender-neutral centre. This was to account for noise in the data and biases arising from the crowdsourcing method. Examples for masculine words include "athletic," "intense," and "tough." Examples for feminine words include "sincere," "cute," and "love."

- Gaucher et al. [48] handcrafted a list for analyzing language use in job advertisements based on explicit (masculine and feminine) and implicit (agentic and communal) associations with various words and gender. Examples of feminine words include "sensitive," "interpersonal," and "communal." Examples of masculine words include "active," "confident," and "decisive."

*Table 1: Dictionaries used in Study 1.1*

| Citation | Year | Gendering | Word Sources | Categorization Method | Format & Interpretation | Total |
|---|---|---|---|---|---|---|
| Roberts & Utych [87] | 2019 | Masculine, neutral, feminine | ANEW [18], curated synonyms | AMT (n=175; 2018) | 1-7-point scale, where fem. <= 3 and masc. >= 5, or | 124 |
| | | | | | fem. < .25 and masc. > 5.5 | 31 |
| Gaucher et al. [48] | 2011 | Masculine, feminine | Agentic and communal words [6,89], masc. and fem. trait words [10,58,93] | Authors (manual) | Categorical list | 79 |

Year: Publication date of the associated paper. Total: Number of tokens after stemming, lemming, and removing stop words.

### 4.1.2 Data Preparation and Analysis.

We used RStudio with R version 4.1.2 (2021-11-01) or "Bird Hippie." We followed the data prep guidelines by Welbers, Van Atteveldt, and Benoit [108]. For the data sets and dictionaries, we used the quanteda default for pattern matching, glob, which uses a combination of the regular expression characters of * for any number of characters and ? for any single character, e.g., masc* and masc?, thereby accounting for most use cases without excluding viable results. We also removed stop words as recommended by the guidelines we followed [108]. We conducted overall analyses but also divided our analysis by ML partition, i.e., train, dev, and test, to isolate the potential impact of bias in the training process. We then followed the automated content analysis procedures by Puschmann and Haim[13] and Hase[14]. We used the quanteda[15] (3.2.0), Tidyverse[16], lexicon, rcompanion, dplyr, car, rstatix, DescTools, and jsonlite R packages. Quanteda is widely-used for natural language processing of text data in qualitative research [108]. It provides packages for conducting automated content analyses of a wide variety of textual data and the key functions for NLP and textual data management. We generated counts, frequencies, and ratios, in line with automated content analysis procedures for NLP data sets [1,1,23,42], as well as other forms of text data [48,87]. We used inferential statistics, such as Chi-square tests, to compare the results of the dictionaries by language gendering. Given the uneven groups, i.e., more feminine words in a dictionary compared to other implicitly gendered words, we used relative frequencies for each set of words given the total in the dictionary as well as corrective statistics, such as the Games-Howell.

---

[13] https://content-analysis-with-r.com
[14] https://bookdown.org/valerie_hase/TextasData_HS2021
[15] https://quanteda.io
[16] https://www.tidyverse.org

For the contextual analysis, we drew on summative content analysis methodology, a manual approach to analyzing qualitative data that is comprised of two phases [60]. Automated content analysis effectively maps onto the first phase: manifest content analysis, wherein counts and sums are produced. Contextual analysis thus represents the next phase: a latent content analysis of these quantitative results within the data sets, where the aim is to interpret these "gendered" patterns against the VA and/or NLP context. This involved multiple authors, working alone or in concert, determining the meanings, potentially multiple, of each word from gender- and context-oriented perspectives by looking at the data sets directly, i.e., how the words were actually used in the utterances, sentences, and dialogues within MASSIVE and ReDial. Disagreements were discussed until consensus was reached.

## 4.2   STUDY 1.1: INITIAL RESULTS

We summarize the results on implicit gendered language for MASSIVE in Table 2 and Figure 1 and ReDial in Table 3 and Figure 2. We found that .23% of MASSIVE and 1.2% ReDial represented implicit gendered language. One-quarter of the dictionary terms were found in MASSIVE and one-half of the dictionary terms were found in ReDial. However, we also found evidence of *ambiguous terms*: words that could relate to gender or be perceived as gendered, but also relate to the VA context and/or technology more broadly (refer to 4.3). Re-conducting our analyses without these words, we found that gendered language was still present, albeit significantly reduced: .08% for MASSIVE and .19% for ReDial. Moreover, there was still a masculine gendered language bias in both data sets, and this was particularly pronounced for MASSIVE. We start by presenting the results on the most frequent words, then compare masculine and feminine gendered language rates, and then our process of identifying ambiguous words given the differences between the data sets and dictionaries. We then report on our analyses without ambiguous terms and compare the data sets directly.

*Table 2: Summary of results for gendered language in the MASSIVE data set (Study 1)*

| Dictionary | Dict. Share[d] | Total Instances | Partition | Masc. Freq. | Fem. Freq. | Masc. Ratio | Fem. Ratio |
|---|---|---|---|---|---|---|---|
| Roberts & Utych [87]: Loose criteria[a] | 38/124 (29.7%) | 172 | overall | 93 | 79 | .541 | .459 |
| | | | dev | 15 | 7 | .682 | .318 |
| | | | test | 14 | 19 | .424 | .576 |
| | | | train | 64 | 53 | .547 | .453 |
| Roberts & Utych [87]: Conservative criteria[b] | 5/31 (22.6%) | 12 | overall | 9 | 3 | .75 | .25 |
| | | | dev | 1 | 0 | 1 | 0 |
| | | | test | 0 | 1 | 0 | 1 |
| | | | train | 8 | 2 | .800 | .200 |
| Gaucher et al. [48] | 14/79 (17.3%) | 87 | overall | 74 | 13 | .149 | .851 |
| | | | dev | 1 | 6 | .143 | .857 |
| | | | test | 3 | 12 | .200 | .800 |
| | | | train | 9 | 56 | .138 | .862 |
| Roberts & Utych [87] and Gaucher et al. [48] combined | 51/209 (24.4%) | 258 | overall | 106 | 152 | .411 | .589 |
| | | | dev | 16 | 13 | .552 | .448 |
| | | | test | 17 | 31 | .354 | .646 |
| | | | train | 73 | 108 | .403 | .597 |
| Combined without AVA[c] terms | 22/169 (13%) | 66 | overall | 54 | 12 | .818 | .182 |
| | | | dev | 9 | 0 | 1 | 0 |
| | | | test | 9 | 2 | .818 | .182 |
| | | | train | 36 | 10 | .783 | .217 |

Dict.: Dictionary. Masc.: Masculine. Fem.: Feminine. Freq.: Frequency. [a]Loose: fem. <= 3 and masc. >= 5. [b]Conservative: fem. < .25 and masc. > 5.5. [c]AVA terms: Ambiguous Virtual Assistant terms. [d]Dictionary share: How many of the total possible dictionary terms were found in the data set, with at least one instance per term.

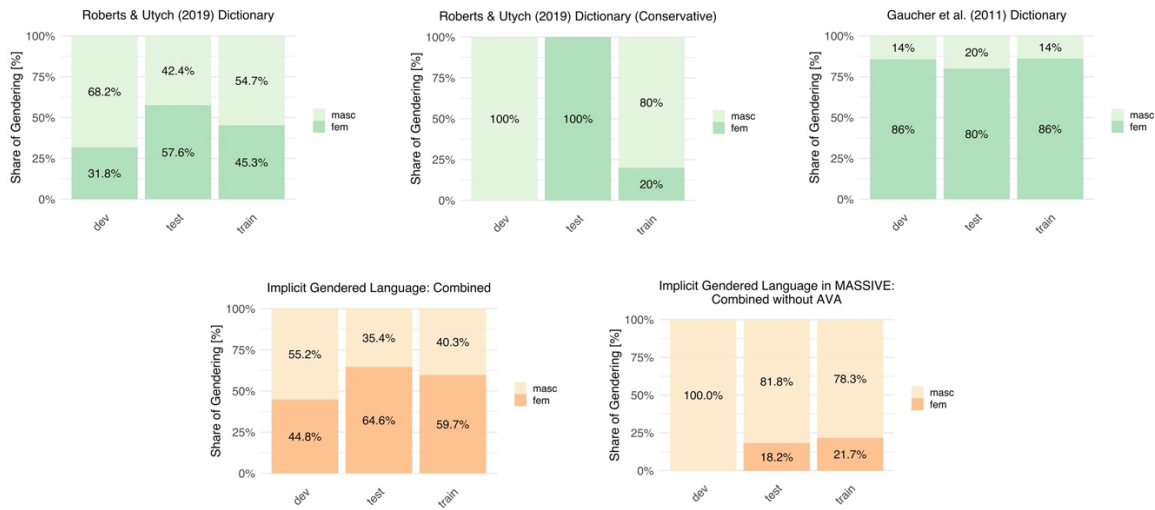

*Figure 1: Visualized results for implicit gendered language in the MASSIVE data set (Study 1).*

Table 3: Summary of results for implicit gendered language in the ReDial data set (Study 1)

| Dictionary | Dict. Share[d] | Total Instances | Partition | Masc. Freq. | Fem. Freq. | Masc. Ratio | Fem. Ratio |
|---|---|---|---|---|---|---|---|
| Roberts & Utych [87]: Loose criteria[a] | 79/124 (61.7%) | 12,420 | overall | 1696 | 10,724 | .137 | .863 |
|  |  |  | test | 162 | 1185 | .120 | .880 |
|  |  |  | train | 1534 | 9539 | .139 | .861 |
| Roberts & Utych [87]: Conservative criteria[b] | 22/31 (71%) | 1820 | overall | 989 | 831 | .543 | .457 |
|  |  |  | test | 109 | 45 | .708 | .292 |
|  |  |  | train | 880 | 786 | .528 | .472 |
| Gaucher et al. [48] | 37/79 (45.7%) | 4328 | overall | 4237 | 91 | .979 | .021 |
|  |  |  | test | 5 | 524 | .009 | .991 |
|  |  |  | train | 86 | 3713 | .023 | .977 |
| Roberts & Utych [87] and Gaucher et al. [48] combined | 113/209 (54%) | 16,697 | overall | 1787 | 14,910 | .107 | .893 |
|  |  |  | test | 167 | 1705 | .089 | .911 |
|  |  |  | train | 1620 | 13,205 | .109 | .891 |
| Combined without AVA[c] terms | 74/169 (43.8%) | 2472 | overall | 1344 | 1128 | .544 | .456 |
|  |  |  | test | 127 | 69 | .648 | .352 |
|  |  |  | train | 1217 | 1059 | .535 | .465 |

Dict.: Dictionary. Masc.: Masculine. Fem.: Feminine. Freq.: Frequency. [a]Loose: fem. <= 3 and masc. >= 5. [b]Conservative: fem. < .25 and masc. > 5.5. [c]AVA terms: Ambiguous Virtual Assistant terms. [d]Dictionary share: How many of the total possible dictionary terms were found in the data set, with at least one instance per term.

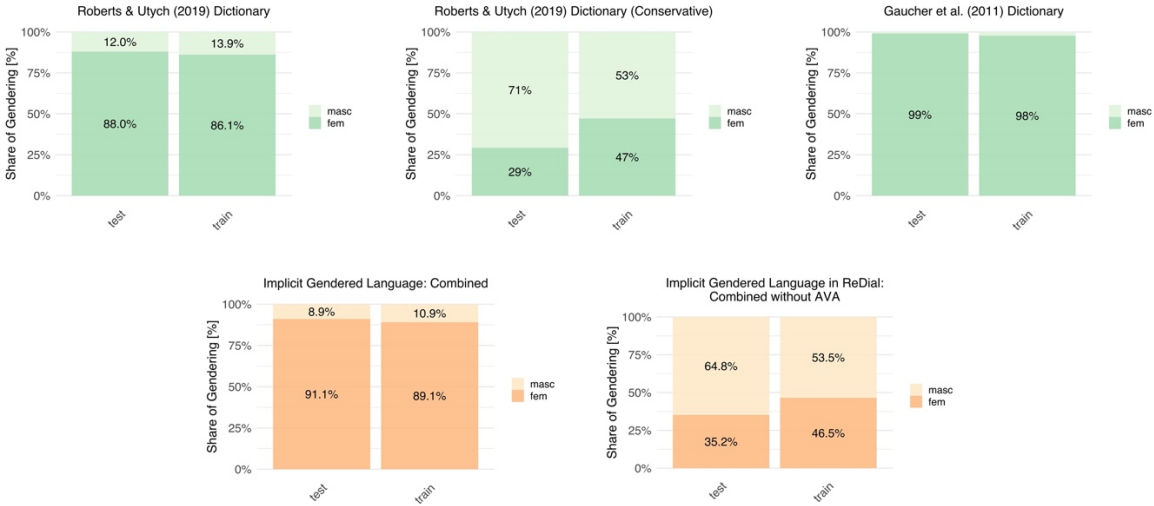

*Figure 2: Visualized results for implicit gendered language in the ReDial data set (Study 1).*

### 4.2.1 Word Shares and Frequencies.

We considered word shares (how many gendered words) and frequencies (how many instances of each gendered word). The results for the 20 most common words per dictionary are in Table 4; the full list is in Appendix A.

*Table 4: Word frequencies for the top 20[a] words in the MASSIVE and ReDial datasets (Study 1)*

| Roberts and Utych [87] | | | | | | Roberts and Utych [87]: Conservative | | | | | | Gaucher et al. [48] | | | | | |
|---|---|---|---|---|---|---|---|---|---|---|---|---|---|---|---|---|---|
| **MASSIVE** | | | **ReDial** | | | **MASSIVE** | | | **ReDial** | | | **MASSIVE** | | | **ReDial** | | |
| Token | G. | Freq. | Token | G. | Freq. | Token | G. | Freq. | Token | G. | Freq. | Token | G. | Freq. | Token | G. | Freq. |
| love | f | 45 | love | f | 642 | man | m | 3 | cute | f | 553 | quiet | f | 17 | kind | f | 380 |
| dog | m | 25 | hope | f | 3028 | guy | m | 3 | man | m | 365 | warm | f | 17 | together | f | 1 |
| power | m | 22 | cute | f | 553 | woman | f | 2 | guy | m | 300 | kind | f | 14 | child | m | 135 |
| soft | f | 8 | man | m | 365 | terrorist | m | 1 | hero | m | 197 | together | f | 13 | honest | f | 79 |
| bake | f | 7 | guy | m | 300 | beautiful | f | 1 | woman | f | 141 | active | m | 7 | emotional | f | 65 |
| direct | m | 5 | hero | m | 197 | captain | m | 1 | beautiful | f | 102 | cheer | f | 4 | trust | f | 49 |
| hope | f | 5 | cry | f | 159 | | | | violent | m | 47 | child | m | 4 | decide | m | 33 |
| teacher | f | 5 | intense | f | 157 | | | | adorable | f | 24 | pleasant | f | 3 | quiet | f | 30 |
| man | m | 3 | woman | f | 141 | | | | prison | m | 23 | analyze | m | 3 | warm | f | 14 |
| grill | f | 3 | director | f | 132 | | | | captain | m | 17 | connection | f | 1 | independent | m | 14 |
| style | m | 3 | | | 115 | | | | rough | | 8 | challenge | f | 1 | understandable | f | 10 |
| | | | | | | | | | | | | | | | | | 9 |

*Table 4: Word frequencies for the top 20[a] words in the MASSIVE and ReDial datasets (Study 1)*

| Roberts and Utych [87] | | | | | | Roberts and Utych [87]: Conservative | | | | | | Gaucher et al. [48] | | | | | |
|---|---|---|---|---|---|---|---|---|---|---|---|---|---|---|---|---|---|
| **MASSIVE** | | | **ReDial** | | | **MASSIVE** | | | **ReDial** | | | **MASSIVE** | | | **ReDial** | | |
| **Token** | **G.** | **Freq.** | **Token** | **G.** | **Freq.** | **Token** | **G.** | **Freq.** | **Token** | **G.** | **Freq.** | **Token** | **G.** | **Freq.** | **Token** | **G.** | **Freq.** |
| love | f | 45 | love | f | 6462 | man | m | 3 | cute | f | 553 | quiet | f | 17 | kind | f | 3801 |
| dog | m | 25 | hope | f | 3028 | guy | m | 3 | man | m | 365 | warm | f | 17 | together | f | 135 |
| power | m | 22 | cute | f | 553 | woman | f | 2 | guy | m | 300 | kind | f | 14 | child | f | 79 |
| soft | f | 8 | man | m | 365 | terrorist | m | 1 | hero | m | 197 | together | f | 13 | honest | f | 65 |
| bake | f | 7 | guy | m | 300 | beautiful | f | 1 | woman | f | 141 | active | m | 7 | emotional | f | 49 |
| direct | m | 5 | hero | m | 197 | captain | m | 1 | beautiful | f | 102 | cheer | f | 4 | trust | f | 33 |
| hope | f | 5 | cry | f | 159 | handsome | m | 1 | violent | m | 47 | child | f | 4 | decide | m | 30 |
| teacher | m | 5 | intense | m | 157 | | | | adorable | f | 24 | pleasant | f | 3 | quiet | f | 14 |
| man | m | 3 | woman | f | 141 | | | | prison | m | 23 | analyze | m | 3 | warm | f | 14 |
| grill | m | 3 | director | m | 132 | | | | captain | m | 17 | connection | f | 1 | independent | m | 13 |
| style | f | 3 | style | f | 115 | | | | rough | m | 16 | challenge | m | 1 | understandable | f | 10 |
| barbecue | m | 3 | beautiful | f | 102 | | | | terrorist | m | 8 | compassionate | f | 1 | connection | f | 9 |
| guy | m | 3 | dog | m | 67 | | | | handsome | m | 6 | intellectual | m | 1 | pleasant | f | 9 |
| virgin | f | 2 | strong | m | 61 | | | | destruction | m | 6 | leader | m | 1 | intellectual | m | 8 |
| crude | m | 2 | buddy | m | 51 | | | | heroine | f | 4 | | | | adventurous | m | 7 |
| rebel | m | 2 | emotional | f | 49 | | | | sassy | f | 4 | | | | commit | f | 5 |
| strong | m | 2 | violent | m | 47 | | | | cherish | f | 2 | | | | challenge | m | 5 |
| director | m | 2 | tough | m | 44 | | | | brash | m | 1 | | | | compete | m | 5 |
| industry | m | 2 | fabulous | f | 27 | | | | jock | m | 1 | | | | cheer | f | 5 |
| tough | m | 2 | adorable | f | 24 | | | | sensitive | f | 1 | | | | leader | m | 5 |

G.: Group. Freq.: Frequency. [a]Up to 20 if that amount exists.

For MASSIVE, the conservative automated content analyses for Roberts and Utych [87] indicated a masculine language bias in word shares (25 masc., 13 fem., 65.8% masc.; 5 masc., 2 fem., 71.4% masc.) and frequencies (93 masc., 79 fem., 54.1% masc.; 19 masc., 13 fem., 75% masc.). In contrast, the results for Gaucher et al. [48] indicated a feminine language bias in word shares (5 masc., 9 fem., 35.7% masc.) and frequencies (13 masc., 74 fem., 14.9% masc.). When applying the loose Roberts and Utych [87] dictionary to ReDial, an implicit masculine language bias was found in terms of shares (52 masc., 27 fem., 65.8% masc.) and an implicit feminine language bias in terms of frequencies (1696 masc., 10,724 fem., 13.7% masc.). For the conservative version, there was an implicit masculine bias in terms of shares (14 masc., 8 fem., 63.6% masc.) but almost a balance in frequencies (989 masc., 831 fem., 54.3% masc.). The Gaucher et al. [48] dictionary indicated a balance in shares (18 masc., 19 fem., 48.6% masc.) but an implicit feminine bias in frequencies (91 masc., 4237 fem., 2.1% masc.). Overall, we discovered apparent disagreement between the dictionaries on the relative masculine and feminine language bias in the data sets. This led us to pause and turn to critically examining why through contextual analysis.

### 4.2.2 *Contextual Analysis.*

Our findings on the most frequent dictionary terms found in each data set (Table 4) highlighted a potential disconnect between the *context* of each *dictionary* in relation to the *context* of each *data set* [53,108]. As Hase, referring to Grimmer and Stewart [53], recommends, we need to assess how the dictionary was constructed, when, and for what situations, as well as what relationships between dictionary terms and words in the data sets would theoretically (or logically) be expected. We also needed to consider the contexts in which the terms were used in the data sets, i.e., VAs and NLP.

For both data sets, the Gaucher et al. [48] dictionary indicated a clear feminine bias. Yet, it has a few special properties that could explain this difference in the context of an NLP data set for VAs and other dialogue-based agents. It is an older dictionary (2011) with few terms (n=82). It was handcrafted by a small group of people (the authors of the paper) who wove together lists of words from various sources for the context of job ads, taking heed of the possible occupations and their gendered associations. This context and the relative frequency of words used within it may not necessarily translate to conversational data sets constructed within daily home and office contexts or in conversations about movies and technology. Moreover, over time, shifts in attitudes towards occupational gender stereotypes may have occurred, in favour of women but not men [110]. Trends indicate that greater rates of women have been taking on traditionally masculine occupations *and* are encouraged to do so, but neither tends to be the case for men when it comes to traditionally feminine occupations. For example, we have bootcamps for girls in computer science, but we do not have bootcamps for boys in nursing and caregiving [84]. A nuanced review of the dictionary terms was needed.

Table 4 in conjunction with a direct examination of the data sets points to instances that may not be about gender, per se. "Love" (MASSIVE n=45, ReDial n=6462), the most frequent word from the Roberts and Utych [87] dictionary, was mainly used to indicate preference, e.g., "I do love scary movies." Other words appeared to be closely related to VA tasks and contexts of use. In the Gaucher et al. [48] dictionary, "warm" (MASSIVE n=17, ReDial n=14), for example, was almost exclusively used in the context of asking about the weather, one of the most typical VA commands identified [105]. "Kind" (MASSIVE n=14, ReDial n=3801) was almost exclusively used in both data sets as a colloquial analogy of "type" or "variety," e.g., "what kind of music?" Still others were multifaceted and contextual. One example from the Roberts and Utych [87] dictionary was "power," which was used in MASSIVE to refer to electricity, e.g., "save power," but in ReDial as a possibly masculine descriptive characteristic, e.g., "all-powerful Hela." In short, a critical lens revealed shortcomings related to contextual differences. To move forward, we needed to address these shortcomings.

## 5 Interlude: Developing a Context-Sensitive Dictionary of Ambiguous VA Terms, or AVA

We did not have confidence in the Study 1.1 results. As such, we evaluated the terms that were found in both data sets—as an indicator of their prevalence—from a critical, reflective angle [88] and generate new materials to enable a more nuanced, contextualized approach to analysis [12,78]. This led us to create a new dictionary of ambiguous terms for the context of VA-oriented data sets: the *Ambiguity for Virtual Assistants* dictionary or *AVA*. We defined *ambiguous terms* as those deemed by existing dictionaries to be instances of implicit gendered language that also relate to the VA context in gendered *or* gender-neutral ways, i.e., there could be multiple legitimate interpretations of a given term that requires contextual analysis to discern. We used the following criteria to code the terms:

- The term was from an implicit gendered language dictionary, i.e., Roberts and Utych [87] and Gaucher et al. [48], *and*

- At least one instance of the term was found in either MASSIVE or ReDial, *and*
- The term was ambiguous in one of the following ways within the VA context of the NLP data sets:
  - The term did not carry the expected implicit gendered language connotations, i.e., it had a special meaning within the VA context that did not appear to be an instance of implicit gendered language but could still carry gendered associations, e.g., the term 'assist' may be neutral or ambiguous, given the feminization of modern VAs and the 'assistant' role, *or*
- When multiple instances of the term were found:
  - At least one instance carried the expected implicit gendered language interpretation of the term, *and*
  - At least one other instance appeared to relate to the VA context and did not carry the expected implicit gendered language interpretation of the term, even if it could still carry gendered associations, e.g., the term 'strength' is used in the context of volume levels but also to describe 'a man with great physical strength'

In creating AVA, we drew from critical discourse analysis [41], focusing on the relations and murky intersections of gendered and mechanical associations embedded in the words and the contexts in which they are found in the data sets. For this, the first author used the Study 1.1 results as a guide to manually read through the data sets and isolate words that were ambiguous in relation to the context, i.e., possibly (implicitly) gendered, possibly VA-specific, or possibly both. Three authors native or fluent in English then carried out independent content analyses of this list, rating each term as ambiguous or not with an example from either MASSIVE or ReDial. As labellers, we relied on our combined expertise in gender research and technology domains to rate each term; refer to Table 5 for relevant demographics. When disagreements arose, the team discussed each case using examples drawn from the data sets until consensus was reached. A Krippendorff's alpha test was used to estimate the inter-rater reliability among the three of us as raters [56]. All terms achieved $\alpha \geq .988$, indicating high agreement.

*Table 5: Labeller demographics for the AVA dictionary terms*

| Labeller | Technical Domain Knowledge | Gender Domain Knowledge |
|---|---|---|
| First author | Researcher in HCI with a Ph.D. degree in engineering, with over a year of postdoc experience in an ML lab and two decades of industry experience in the tech sector. Have been developing and publishing research on VAs, especially on the topic of speech, for three years, and have | Have been involved in feminist and social justice spaces for a decade and have published feminist work on robots in the last two years. |

| Labeller | Technical Domain Knowledge | Gender Domain Knowledge |
|---|---|---|
| | also been working and publishing in adjacent spaces, including social robotics, for ten years. | |
| Second author | Researcher in HRI with a Ph.D. degree in engineering. Nine years of experience working with social robots and designing verbal and non-verbal behaviours in robots as social agents. Have been working on the topic of speech dysfluency for 3 years. Have conducted several HRI user studies in which social robots use speech as a primary mode of communication. | Have been involved in research on and conducting workshops related to equity, diversity, and inclusion at academic venues in the last two years. |
| Third author | Ph.D. student in HCI with a M.Sc. degree in engineering. Experience with and knowledge in ML with passing knowledge on using ML models for NLP. | Following current LGBTQ+ issues with a vested interest in a more equal and open-minded society. |

From an initial set of 185 terms, we isolated 44 for inclusion, with 26 deemed masculine by the original dictionaries and 18 deemed feminine. Examples are "love" ("I love action movies"), "power" ("powered off"), "connection" ("internet connection"), "kind" ("kind of weather"), and "quiet" ("be quiet"). We formatted the dictionary in JSONL, which includes the term, its genderedness as originally assigned in the dictionary, and examples from the MASSIVE and ReDial data sets showing how it has an ambiguous application in VA contexts. The JSONL is in the Supplementary Materials.

## 6   Study 1.2: Follow-Up Analysis of Gendered Language with AVA

We used the AVA dictionary to remove ambiguous terms from the combined dictionary and then re-ran the previous analyses; refer to Section 4 for details. We used a removal strategy to isolate the genderedness of non-AVA terms, i.e., terms that are not ambiguous in the VA/NLP content, in the data sets. Summaries for MASSIVE are in Table 2 and Figure 1, and for ReDial, refer to Table 3 and Figure 2; the most common words for both are in Table 6.

*Table 6: Word frequencies for the top 20 words in the MASSIVE and ReDial datasets without AVA terms (Study 1.2)*

| MASSIVE | | | ReDial | | |
|---|---|---|---|---|---|
| Token | G. | Freq. | Token | G. | Freq. |
| dog | m | 25 | cute | f | 553 |
| active | m | 7 | man | m | 365 |
| teacher | m | 5 | guy | m | 300 |

| MASSIVE | | | ReDial | | |
|---|---|---|---|---|---|
| Token | G. | Freq. | Token | G. | Freq. |
| man | m | 3 | hero | m | 197 |
| pleasant | f | 3 | cry | f | 159 |
| guy | m | 3 | intense | m | 157 |
| virgin | f | 2 | woman | f | 141 |
| rebel | m | 2 | beautiful | f | 102 |
| buddy | m | 2 | dog | m | 67 |
| woman | f | 2 | buddy | m | 51 |
| peaceful | f | 1 | emotional | f | 49 |
| cry | f | 1 | fabulous | f | 27 |
| abuse | m | 1 | adorable | f | 24 |
| chief | m | 1 | bud | m | 24 |
| compassionate | f | 1 | prison | m | 23 |
| furious | m | 1 | powerful | m | 19 |
| beautiful | f | 1 | captain | m | 17 |
| captain | m | 1 | thoughtful | f | 13 |
| attractive | f | 1 | furious | m | 12 |
| intellectual | m | 1 | teacher | m | 12 |

G.: Group. Freq.: Frequency.

First, we evaluated the relative gender biases in terms of frequencies with AVA terms removed. For MASSIVE, a Chi-square goodness-of-fit test indicated that the ratio of masculine and feminine gendering was inconsistent, with more implicit masculine (n=54) than feminine (n=12) words found, $\chi2 = 26.73$; $df = 1$; $p < .001$. A similar pattern was found for ReDial, with more implicit masculine (n=1344) than feminine (n=1128) words found, $\chi2 = 18.87$; $df =1$; $p < .001$. In short, accounting for ambiguous terms revealed evidence of implicit masculine language biases. We then compared the distributions of implicit masculine and feminine language, without AVA terms, by data set. We found that the relative frequencies significantly differed from expectation, $\chi2 = 18.48$; $df = 1$; $p < .001$, with greater divides in MASSIVE. This indicates that MASSIVE has a greater degree of implicit masculine language bias.

## 6.1   DISCUSSION

Both the MASSIVE and ReDial data sets were found to be implicitly biased with gendered language. This was to a small degree relative to all the words in the data set. We must also consider this against the relatively small size of the dictionaries and the data sets themselves. Yet, we cannot ignore the presence of implicit gendered language. Even when we account for ambiguous cases with the AVA terms, the bias persists, and it is masculine valanced. While we cannot evaluate the effect that this biased language may or may not have, it is troubling, but potentially avoidable.

A key contribution of Study 1 was the AVA terms, which can be used in and outside of this research to isolate words with layered meanings in terms of gender and VA contexts for various purposes, and not only debiasing. The efficacy of AVA deserves a more nuanced discussion. The data sets may be different in important ways. ReDial may contain more implicit gendered language and a stronger masculine bias because it is older, or because it is about movies, which may give rise to discourse that is relatively more gendered and masculinized than the kinds of discourse that may occur in a home or office setting with a daily assistive robot, i.e., MASSIVE. We must also interrogate these findings in light of the gendered associations of these words, who was involved in their creation, i.e., mostly men in positions of power, and the existence of legitimate alternatives [29,107]. Computer technology has largely been a masculinized domain dominated by men [29,107], and this has led to the conscious and unconscious use of value-laden terminology built into tools, such as programming languages and praxis. ACM[17] and others in the computing profession have taken strides to correct this by advocating against sexist, racist, ableist, ageist, and privileged language [70]. We might ask: why light "intensity" rather than "levels" and "power" over "electricity"? Likewise, "quiet" (n=17,15) could be about volume levels *or* rude demands that may have a sexist, toxic element to them. This would not be unexpected because most intelligent agents have stereotypically feminine-gendered, subservient embodiments that are known to elicit gender-based harassment, by naïveté or by design [44,59,100]. We must also consider the gaps. For example, why are there so many mentions of dogs and so few of cats? Recent statistics on pet ownership from the American Veterinary Medical Association[18] would suggest two-fifths or 10 mentions of cats relative to the 27 mentions of dogs. Moreover, half of the mentions to cats were to a person's name rather than the animal. This may relate to long-standing gendered associations of masculinity with dogs and femininity with cats, whereby dogs receive more space and recognition than cats [96]. We must consider the scale and layers as well as the intersections between features of the data sets, language, and society.

---

[17] https://www.acm.org/diversity-inclusion/words-matter
[18] https://www.avma.org/resources-tools/reports-statistics/us-pet-ownership-statistics

Implicit gendered language is but one form of implicit gendering that could be present in NLP data sets. We now turn to another for comprehensiveness and comparison: masculine-as-norm language.

# 7   Study 2: Masculine-as-Norm Language

We next evaluated the relative presence of masculine-as-norm language. We operationalized masculine-as-norm language according to the literature: pronoun use, gender-marked words, and names.

## 7.1   METHODS

We used the same procedure as in Study 1, i.e., automated content analysis, with a different selection of dictionaries. When running our analysis with the pronoun dictionary, we did not remove stop words, which include pronouns.

### 7.1.1   *Dictionaries.*

An overview of the dictionaries we used is presented in Table 7. As before, we chose dictionaries that were relevant, established, and freely available. We also tried to account for the relative strengths and weaknesses of the range of dictionaries available, especially in terms of genders beyond the binary. We describe each dictionary below:

- For pronoun use, we selected the pronouns dictionary by Lauscher, Crowley, and Hovy [71] because it includes a diversity of pronouns outside of the gender binary. It is divided into pronoun categories, including gendered pronouns, e.g., he/her/him, gender-neutral pronouns, e.g., they/them/their, and neopronouns, e.g., ze/zir.[19] It was compiled by the authors from several sources, including academic publications and community texts.

- For gender-marked words, we used the gendered words subset of the Multi-Dimensional Gender Bias Classification dataset created by Dinan et al. [36]. The dictionary is gender binary, only containing gendered words typically used to refer to men and women. The genderedness of the words was annotated through AMT crowdsourcing. Notably, most of the mturkers were men-identifying (67.38%) and few were non-binary (0.21%), although ~14% did not indicate a gender identity. Examples of masculine words include "policeman," "cowboy," and "dude." Examples of feminine words include "policewoman," "cowgirl," and "gal." Note that gender marked words may refer to specific subjects ("Misako, a policewoman") or not ("a policewoman takes charge") [20]; we do not assess this, but rather aim to capture the genderedness of language overall.

---

[19] We also did not use an exhaustive list, as there is no formalized dictionary yet, which is beyond the scope of this work to create; future research may wish to start with community resources such as https://en.pronouns.page/pronouns

- We also used the Gendered Words Dataset[13] for gender-marked words. It was compiled by GitHub contributors ecmonsen, phseiff, and Guy Rapaport (@guy4261), who drew from the WordNet® data set [43]. They manually tagged hyponyms for "person" as masculine, feminine, neutral, or other. Examples of masculine words include "businessman," "codger," and "wingman." Examples of feminine words include "cat," "dame," and "jezebel." Examples of neutral words include "person," "advocate," and "codefendant."

- For names, we used the name genders subset of the Multi-Dimensional Gender Bias Classification dataset created by Dinan et al. [36], which uses the 1879-2019 US census data. Gender neutral names were those attributed to at least one "man" and one "woman" in this sex/gender binary-based census, e.g., Jaime, Riley, Taylor. We recognize that this approach is limited. Names do not necessarily indicate gender or sex [20,34]. Names are typically assigned at birth based on apparent sex, which may not match internal sex characteristics or gender identity. Not everyone may perceive a given name as gender neutral. Even when written in English, names may have origins or reorientations based in religion, ethnicity, cultural background, nationality, and so on that are not acknowledged in normative English contexts. For example, Two-Spirit people of Turtle Island (Indigenous North America) may be assigned a gender identity and name within the Western colonial context. Yet, "Two-Spirit" is pluralistic, varying across Indigenous communities and cultures, intersecting with sex, sexuality, and non-Western cultural factors in ways that do not always map onto the Western-centred "LGBTQI+" model, such that no "translation" to English would be appropriate [32]. We do not know the extent to which any of these patterns are the case for deemed gender neutral or other names; it likely varies across time, context, and individual. Still, given masculine-as-norm biases, gender neutral or ambiguous names may be read as masculine; on the same token, feminine names may not necessarily be read as feminine. In short, names are a source of gender bias, where people make judgments, knowingly or not, about the gender of the person based on their understanding of that name within their particular socio-cultural context [20]. We include names as salient feature in dialogic materials that can act as a gender stimulus.

*Table 7: Dictionaries used in Study 2*

| Citation | Year | Referent Type | Gendering | Word Sources | Categorization Method | Format & Interpretation | Total |
|---|---|---|---|---|---|---|---|
| Lauscher, Crowley, & Hovy [71] | 2022 | Pronouns | Masculine, feminine, neutral, neo[a] | Various | Authors (manual) | Categorical lists | 49 |
| Dinan et al. [36] | 2020 | Marked Words | Masculine, feminine | AMT | Authors (manual) | Categorical lists | 409 |
| Gendered Words Dataset[20] | 2019 | Marked Words | Masculine, feminine, neutral, other | WordNet®[21], hyponyms | Authors (manual | JSON, gender as m, f, n, o | 6244 |
| Dinan et al. [36] | 2020 | Names | Masculine, feminine, neutral | AMT | Authors (manual) | Categorical lists | 22,609 |

Year: Publication date of the associated paper. Total: Number of tokens/words in the dictionary. [a]Note: We do not include nounself, emojiself, numberself, and nameself pronouns due to the difficulty in discerning from non-pronouns and hypothetical use in reality.

## 7.2 RESULTS

The results for MASSIVE are in Table 8 and Figure 3; ReDial is in Table 9 and Figure 4. We start with pronouns, then consider marked words using both dictionaries, and end with names.

*Table 8: Summary of results on masculine-as-norm language for the MASSIVE data set (Study 2)*

| Dictionary | Dict. Share | Total Instances | Partition | Masc. Freq. | Fem. Freq. | Neu. Freq. | Neo. Freq. | Masc. Ratio | Fem. Ratio | Neu. Ratio | Neo. Ratio |
|---|---|---|---|---|---|---|---|---|---|---|---|
| Pronouns | 9/49 (18.4%) | 183 | overall | 71 | 50 | 112 | 8 | .388 | .273 | .612 | .044 |
| | | | dev | 7 | 5 | 16 | 3 | .226 | .161 | .516 | .097 |
| | | | test | 14 | 6 | 27 | 1 | .292 | .125 | .562 | .021 |
| | | | train | 50 | 39 | 69 | 4 | .309 | .241 | .426 | .025 |

---

[20] https://github.com/ecmonsen/gendered_words
[21] https://wordnet.princeton.edu

| Dictionary | Dict. Share | Total Instances | Partition | Masc. Freq. | Fem. Freq. | Neu. Freq. | Neo. Freq. | Masc. Ratio | Fem. Ratio | Neu. Ratio | Neo. Ratio |
|---|---|---|---|---|---|---|---|---|---|---|---|
| Dinan et al. [36] marked words | 73/409 (17.9%) | 428 | overall | 185 | 243 | | | .432 | .568 | | |
| | | | dev | 19 | 31 | | | .380 | .620 | | |
| | | | test | 30 | 36 | | | .455 | .545 | | |
| | | | train | 136 | 176 | | | .436 | .564 | | |
| Gendered Words Dataset[22] | 366/6244 (5.7%) | 2768 | overall | 252 | 195 | 2516 | | .091 | .070 | .909 | |
| | | | dev | 28 | 22 | 320 | | .076 | .059 | .865 | |
| | | | test | 49 | 30 | 411 | | .100 | .061 | .839 | |
| | | | trained | 175 | 143 | 1785 | | .083 | .068 | .849 | |
| Dinan et al. [36] names without agent names[a] | 843/22,609 (3.7%) | 6882 | overall | 3328 | 3554 | 2618 | | .484 | .516 | .38 | |
| | | | dev | 405 | 444 | 329 | | .344 | .377 | .279 | |
| | | | test | 634 | 643 | 460 | | .365 | .370 | .265 | |
| | | | overall | 2289 | 2467 | 1829 | | .348 | .375 | .278 | |

Dict.: Dictionary. Masc.: Masculine. Fem.: Feminine. Neu.: Neutral. Neo.: Neopronoun. Freq.: Frequency. [a]Specifically: Alexa, Siri, Olly.

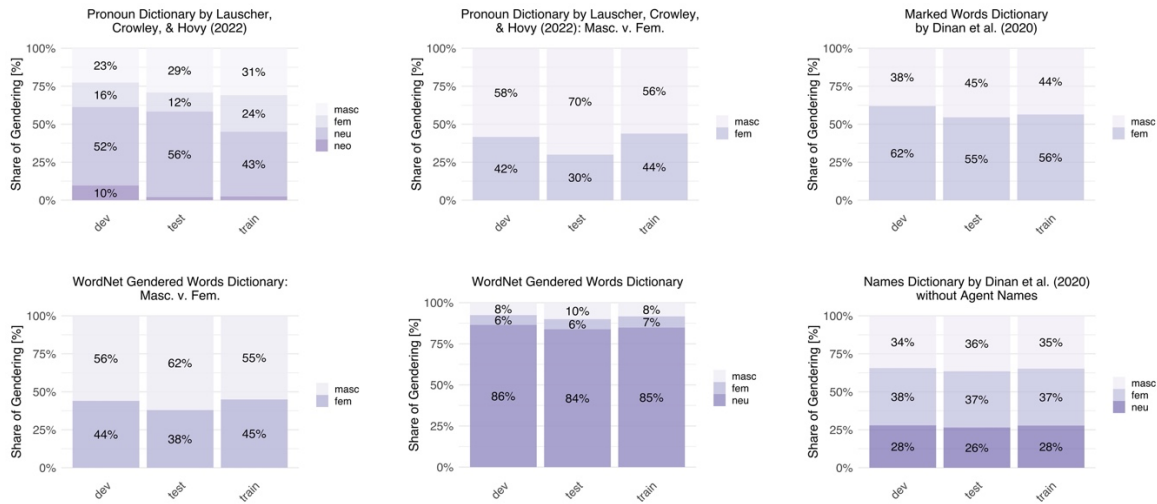

*Figure 3: Visualized results from the masculine-as-norm dictionaries for the MASSIVE data set (Study 2). Note that "neo" stands for neopronouns, e.g., ze/zir. Also note that neopronouns are difficult to automatically detect; counts may be inaccurate.*

*Table 9: Summary of results on masculine-as-norm language for the ReDial data set (Study 2)*

| Dictionary | Dict. Share | Total Instances | Partition | Masc. Freq. | Fem. Freq. | Neu. Freq. | Neo. Freq. | Masc. Ratio | Fem. Ratio | Neu. Ratio | Neo. Ratio |
|---|---|---|---|---|---|---|---|---|---|---|---|
| Pronouns | 19/49 (38.8%) | 9927 | overall | 3501 | 1411 | 6426 | 112 | .353 | .142 | .647 | .011 |
| | | | test | 347 | 105 | 637 | 6 | .317 | .096 | .582 | .005 |
| | | | train | 3154 | 1306 | 5789 | 106 | .305 | .126 | .559 | .010 |
| Dinan et al. [36] marked words | 153/409 (37.4%) | 9845 | overall | 6691 | 3154 | | | .68 | .32 | | |
| | | | test | 657 | 246 | | | .728 | .272 | | |
| | | | train | 6034 | 2908 | | | .675 | .325 | | |
| Gendered Words Dataset | 794/6244 (12.4%) | 38,469 | overall | 6122 | 2840 | 32,347 | | .159 | .074 | .841 | |
| | | | test | 621 | 203 | 3772 | | .135 | .044 | .821 | |
| | | | train | 5501 | 2637 | 28,574 | | .150 | .072 | .778 | |
| Names without agent names[a] | 1541/22,609 (6.8%) | 74,657 | overall | 49,909 | 24,748 | 15,528 | | .669 | .208 | .331 | |
| | | | test | 5632 | 1682 | 2205 | | .574 | .171 | .255 | |
| | | | train | 44,277 | 13,846 | 22,243 | | .552 | .172 | .277 | |

Dict.: Dictionary. Masc.: Masculine. Fem.: Feminine. Neu.: Neutral. Neo.: Neopronoun. Freq.: Frequency. [a]Specifically: Alexa, Siri, Olly. Note: No results for other pronouns.

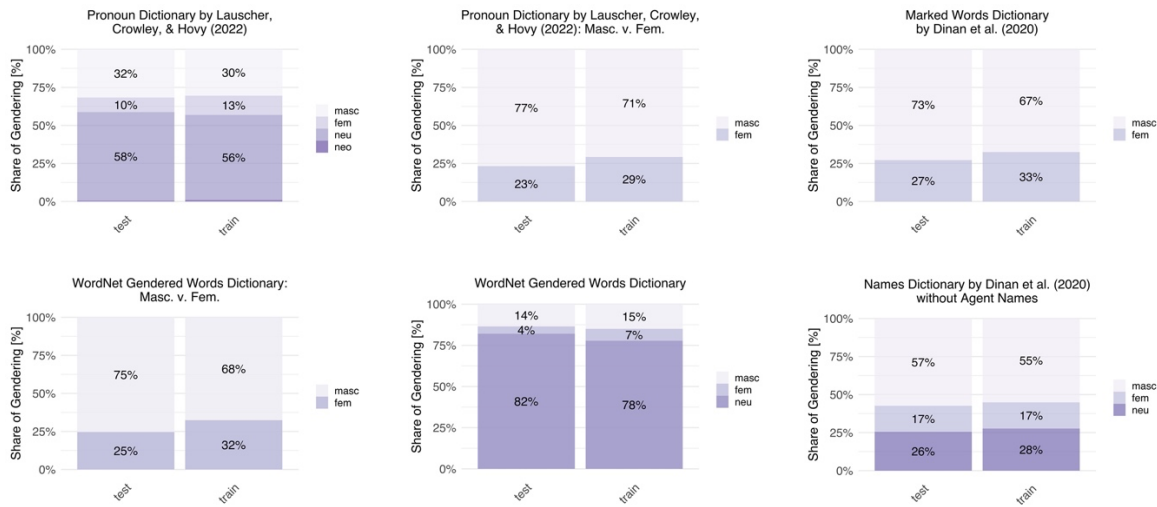

*Figure 4: Visualized results from the masculine-as-norm dictionaries for the ReDial data set (Study 2). Note that "neo" stands for neopronouns, e.g., ze/zir. Also note that neopronouns are difficult to automatically detect; counts may be inaccurate.*

### 7.2.1 Pronouns.

Most pronouns were plural and/or gender neutral, e.g., they/them. This is a positive sign if we are aiming for gender neutrality and inclusivity. However, it is difficult to determine when they/them pronouns are used in the singular, when they are used to refer to multiple entities, and when they are used in an objective sense. For example, one utterance states "who emailed me recently and what did they [ambiguous singular or plural] email about," while another states "check my new emails and tell me what are they [the emails] about" (our emphasis). Moreover, we are not confident in the neopronoun results, as manual checks indicate that many are abbreviations of other pronouns ("them" → "em") or typos. Those results should be interpreted with caution. We begin with MASSIVE. A Chi-Square test did not find a significant difference in the distributions of the masculine (n=71), and feminine (n=50) groups from expected, $\chi^2 = 3.65$; $df = 1$; $p = .056$. This borderline significance may indicate a small masculine pronoun bias with larger samples. However, a Chi-Square tests found significant differences between masculine (n=71), $\chi^2 = 9.19$; $df = 1$; $p = .002$, and feminine (n=50), $\chi^2 = 23.73$; $df = 1$; $p < .001$, groups compared to the neutral (n=112) group. This indicates that there were far more neutral or gender-ambiguous pronouns in MASSIVE. For ReDial, masculine (n=3501) and feminine (n=1411) groups significantly differed from the expected distribution, $\chi^2 = 889.27$; $df = 1$; $p < .001$, indicating a masculine pronoun bias. However, Chi-Square tests found that there were significantly more gender neutral or ambiguous pronouns (n=6426) compared to the masculine, $\chi^2 = 861.85$; $df = 1$; $p < .001$, and feminine, $\chi^2 = 3209.2$; $df = 1$; $p < .001$, groups. This suggests that, like MASSIVE, ReDial has more instances of gender neutral or ambiguous pronouns.

### 7.2.2 Marked Words.

We start with Dinan et al. [36]. For MASSIVE, a significant difference in the expected distributions for masculine (N=185) and feminine (N=243) marked words was found, $\chi^2 = 7.86$; $df = 1$; $p = .005$, indicating a feminine bias. For ReDial, the opposite pattern was found for masculine (n=6693) and feminine (n=3154) groups, $\chi^2 = 1271.9$; $df = 1$; $p < .001$. In short, MASSIVE was feminine-biased and ReDial was masculine-biased. Taking a critical look at the frequencies of individual words in the dictionary indicates a subtle masculine bias. The most frequent marked words for MASSIVE were: mom (n=63), wife (n=15), mother (n=15), lady (n=13), sister (n=12), dad (n=12), king (n=10), papa (n=9), and girlfriend (n=8). We can consider the influence of the dominate gender-sexuality model in most societies today, including the US:

cisgendered and heterosexual, or cishet. A view from this lens suggests that these results are not unexpected. Mothers are the primary caregivers in many families. While not necessarily the case, women are stereotypically thought to engage in more social dialogue than men [65]. We do not have access to the full demographics, including sexualities, of the data set contributors [7]. Nevertheless, these patterns speak to a cishet framing of gendered associations in US society that anyone, including women, may unconsciously embed into dialogues. In contrast, the marked words for ReDial are broader and appeared to be tied to the movie recommendation context: setting aside pronouns, the most frequent words included actor (n=444), man (n=365), actors (n=305), guy (n=300), girl (n=229), daughter (n=207), hero (n=197), husband (n=191), and king (n=163). We can tease out subtle biases and sexism embedded in these terms. We have men and husbands and kings, and we have girls. "Girls" could be contextualized as colloquial expressions, queer language, or the typical infantizliation of women [75]. "Guy" and potentially "actor/s" and "hero" may be more nuanced; while masculine, they are also colloquially used in gender neutral ways, depending on the context. This is, of course, still an expression of masculine-as-norm language: masculine associations raise value, while feminine associations decrease value [84]. As for "king," most references were to famous restaurants, locations, and people. In short, "feminine" trends may actually point to a subtle form of masculine bias in line with masculine-as-norm language.

We then used the Gendered Words Dataset on both data sets. For MASSIVE, there were more masculine (n=256) than feminine (n=209) marked words, $\chi2$ = 4.75; $df$ = 1; $p$ = .03. However, most were neutral (n=2538), as shown against the masculine, $\chi2$ = 1863.8; $df$ = 1; $p$ < .001, and feminine, $\chi2$ = 1974.6; $df$ = 1; $p$ < .001, distributions. The same was true in ReDial for the masculine (n=6180) and feminine (n=2875) distributions, $\chi2$ = 1206.3; $df$ = 1; $p$ < .001, as well as when comparing the neutral (n=32,562) distribution to the masculine, $\chi2$ = 17965; $df$ = 1; $p$ < .001, and feminine, $\chi2$ = 24870; $df$ = 1; $p$ < .001. Looking at the most frequent words aside from pronouns, we can see similar patterns as with the Dinan et al. [36] dictionary. For MASSIVE, these included amazon (n=33), tom (n=32), town (n=26), dog (n=25), wife (n=15), mother (n=15), lady (n=13), sister (n=12), dad (n=12), page (n=11), and king (n=10). For ReDial, these included tom (n=439), man (n=365), guy (n=300), girl (n=229), daughter (n=207), husband (n=191), king (n=163), boy (n=152), son (n=143), and woman (n=141). "Tom" was not found in the analysis of marked words from the Dinan et al. [36] dictionary and seems to refer to user contacts ("Tom" is a common name) and celebrities. We should also wonder why this might be. Why not "Jane," as well? Why are certain names (of people or products) included to a greater degree over others?

A more balanced data set would include a balanced proportion of common names. We take a deeper look at names next.

### 7.2.3 Names.

The names dictionary indicated gender biases and also large shares of gender-neutral and gender-ambiguous names. For MASSIVE, the distribution of masculine (n=3328) and feminine (n=3554) names differed from expectation, favouring the feminine, $\chi2$ = 7.4; *df* = 1; *p* = .006. Masculine, $\chi2$ = 848; *df* = 1; *p* < .001, and feminine, $\chi2$ = 142; *df* = 1; *p* < .001, distributions also differed from the gender-neutral or gender-ambiguous names distribution (n=2618), as well. This indicates a feminine bias. For ReDial, the masculine distribution (n=49,909) was greater than the feminine one (n=15,528), $\chi2$ = 18064; *df* = 1; *p* < .001, and the neutral/ambiguous one (n=24,748), $\chi2$ = 8479.8; *df* = 1; *p* < .001. However, the feminine distribution was much lower than the neutral/ambiguous one, $\chi2$ = 2110.6; *df* = 1; *p* < .001. This indicates a masculine bias. Taken together, each data set appears to have a gender bias, with MASSIVE containing more feminine names and ReDial containing more masculine ones. Yet, each also contains a significant portion of gender neutral or ambiguous names. Many names appear to be atypical *as names* but typical as other kinds of words. For example, "Great" is a masculine name that is also a very common adjective (MASSIVE, n=19; ReDial, n=14,330). A gender neutral or ambiguous example is the name "Awesome," which is also a common word (ReDial, n=1604). Nevertheless, these words carry gendered and non-gendered meanings that are also contextual. We can think of these names as another layer in our approach to examining the implicit gendered language that exists in these data sets.

## 7.3 DISCUSSION

We found an overall implicit cishet masculine bias when contextualizing frequent terms, including pronouns, marked language, and names. A masculine bias was more pronounced in ReDial than MASSIVE. ReDial contained more masculine terms overall. However, while MASSIVE contained more feminine words, interrogating these words revealed a relationship to cishet norms rather than expressions of femininity in its own right, detached from masculinity. Pronouns and names indicated a substantial pattern of potential gender neutrality and gender ambiguity. Nevertheless, it is difficult to tease out whether pronouns are singular or plural, whether names were meant in gendered ways or not, and whether or not this matters in the end. These results may reflect recent efforts to eliminate gender bias by targeting conspicuous patterns like pronouns and name use. We also cannot ignore the presence of the other more subtle masculine bias results alongside this more "balanced" patterns. While encouraging, these results highlight that we still have some way to go in recognizing and accounting for a range of gender biases in NLP data sets.

## 8    Discussion Synthesis

We were able to detect subtle forms of gender bias in the MASSIVE and ReDial data sets, identifying feminine and masculine biases, but more of the latter. Gender neutral and ambiguous language as also identified as a key feature. Moreover, language related to the VA context intersected with "gendered" language in important ways. This led us to change our approach, take on a more critical frame, and accept that our analyses would be limited by this complexity. We submit this work as a first start. While challenging and inexact at times, our approach has set the stage for future efforts on a pressing issue. In particular, we offer an initial dictionary—the AVA terms—as a springboard for more rigorous analysis. We now discuss the main challenges and implications that this initial effort has revealed.

### 8.1    INTERROGATING SUBTLE MASCULINE CENTRISM (AND OTHER) BIASES

Evaluating large NLP data sets using automated procedures is tricky. Language is dynamic, multi-faceted, and contextual. Slang and colloquialisms, names that are also common words, words that are gender ambiguous, and the influence of context; these are just some of the challenges we encountered when attempting to make sense of the more implicit forms of gender bias that may exist in these crowdsourced data sets. One key issue going forward is distinguishing singular "they/them" pronouns from the plural "they/them," ideally in a way that allows for automated content analysis. We should also be wary of implicit feminine language biases. While the context of these words matters, so does the frequency [39]: the more we come across a word, the more we key into it. Even if, for example, we use gender debiasing measures to create feminine associations with masculine words, we should be careful to avoid introducing a feminine gender bias.

## 8.2 A MATTER OF PERSPECTIVE? MANAGING DISAGREEMENTS AMONG THE DICTIONARIES

Dictionaries created for a particular context are highly dependent: not only on the context, but also on the language, culture, local community, religion, race/ethnicity, sexuality, and, of course, sex/gender of the people whose data is being used in the creation process, as well as who the creators are. A dictionary (e.g., for job ads) may not apply to all data sets (e.g., VA data sets). The dictionary share percentages for MASSIVE and ReDial (Table 2 and 3) show how results can differ significantly across different dictionaries and data sets. Due to such incongruities, many dictionary terms may not be found in a given data set, further limiting the efficacy of such analyses. Additionally, due to differences in context, dictionary words may carry unintentional meanings, resulting in lexically or syntactically ambiguous terms that may not be associated with gender in the expected ways, or at all. For instance, some "feminine" words, such as "soft," "kind," and "warmth," were not used in a gendered way in the MASSIVE dataset. At the same time, we must be wary of how we gender VAs, including through the language we train them on [44,100]. These results indicate a need for special dictionaries and word embeddings; indeed, we offer the AVA terms as a starter kit.

## 8.3 MASCULINITIES AND DEEP (UN)LEARNING

We approached language as a medium of power that "engenders" values and ideas. As such, language can also be consciously harnessed to encode desires and possibilities on the cusp of our collective imagination. Considering the results of our both studies, there appears to be masculine-centric language and masculine-as-norm biases in the MASSIVE and ReDial datasets. We must bear in mind the impact of using such data sets at a large scale across a variety of VAs, social robots, and other NLP-based technologies. We must recognize the power of our imagination when it comes to "reshaping technologies" as "socially agentic" and the symbiotic relationship we have with such technologies, which potentially learn from us [80]. We are already grappling with the problem of most VAs defaulting to feminine-sounding voices paired with polite and deprecating behaviour, summarized in the quotation shared across the tiles of a UNESCO 2019 report and Bergen's [13] article on the history of sexism and cyborgs: "I'd Blush If I Could" [109]. These defaults propagate harmful gender biases by reinforcing that feminine-gendered agents should be in subservient roles and tolerate poor treatment. This is not just on the VA side; it is also on the side of the user, who can issue "a blunt voice command like

'hey' or 'OK'" [109, p. 150]. We rely on resources such as open data sets to train our VAs. In doing so, we encode all of the biases in these data sets, blatant and subtle, within our VAs. Recycling existing gendered biased language is thus an enormous possibility. If these biases are not addressed now, they will proliferate.

Harms may be immediate and representational, especially through the embedding of limited and stereotyped representations; they may also be allocational, driving more distant effects that go beyond language [14]. The implications may only arise or become clear over time, especially as we continue to take up these devices at younger and younger ages [38]. For example, speaking with VAs like Alexa may lead to *speech alignment*, a behaviour where speakers adopt the voice and language patterns of other parties unconsciously [112]. This raises several questions: When Alexa speaks with our children, what messages do we want them to take in? Do we want to present a limited version of the world, one wherein masculinity abounds in language use and representation? Do we want Alexa to speak in gender-limited ways? Do we wish to encode "Alexa" as feminine and its masculine-voiced counterpart(s) as masculine through gendered language as well as vocal gender qualities? If children speak in masculine-as-norm ways, how should VAs respond to them? Do we need to create a new "ideal responses" data set to tackle this, or is it okay? We do not and cannot prescribe a one-size-fits all approach. Rather, we raise these provocations for the community to ponder.

We also identified major limitations that may apply to NLP data sets for VAs generally. While both data sets were relatively recent, large-scale, and diverse, with MASSIVE offering a range of languages, they were limited in several ways. Both were premised in basic commands and dialogues representing (stereo)typical VA usage. MASSIVE, for instance, was initiated based on 200 predefined prompts. This leads to a limited range of content. Both also did not include any words related to swearing, politics, sex, drug use, deviancy, sexuality, war, etc. However, people will ask such questions to VAs. This limits practical and real-life applications. Finally, as the data was gathered using observational research methods and crowdsourcing on AMT, just taking part in the studies and being conscious of being observed and recorded likely changed people's natural behaviours, biasing their verbal, non-verbal, and/or written responses [69]. We may need to distinguish between data sets that train VAs how *to speak* and data sets that train VAs on how they may be *spoken to*. Although efforts have been made to mitigate these biases using large and/or outdated corpuses, new challenges arise when the data sets are limited in terms of diversity or size, as well as due to the complexity of how gender frameworks are embedded in language and data sets [72]. We may need to develop more "meta dictionaries" like AVA that map existing corpora to gender-sensitive models of VA and user behaviour.

### 8.3.1 AVA: Ambiguity in Virtual Assistant Language Dictionary

Ambiguity in language is ubiquitous; it is not merely the absence of clarity but also the presence of different meanings that are disambiguated. During this work, we created AVA, a dictionary of 44 VA-specific ambiguous words that were originally classified as gendered by two implicit gendered language dictionaries but became ambiguous against the VA contexts of the MASSIVE and ReDial data sets. Examples include "strong" ("how strong the dollar is compared to the peso" or "make a strong coffee"), "power" ("I have always loved the idea of having some sort of power" or "power up the plug socket"), "soft" ("I have a soft spot for kung fu movie" or "is soft cheese better than hard" or "is it a soft evening tonight"), and "warm" ("is it warm outside" or "to warm my heart"). When used in the VA context, these words can be associated with a certain gender, or not. The creation of the AVA dictionary is a first and crucial step towards identifying and analyzing such ambiguous, multilayered, implicit gender biases in VA data sets. It follows in the footsteps of previous efforts on data statements [12] and model cards [78] by recognizing (a lack of) diversity in NLP initiatives. Further steps include enriching the AVA dictionary by adding more VA-related words representing diverse tasks, commands, and intents, as well as further investigating if these deemed gendered words remain so in VA contexts through controlled empirical studies with gender-diverse people. Indeed, contributions from and co-design activities with genderqueer, non-binary, and trans* folk could help AVA become a truly gender-diverse and -inclusive baseline tool without the use of sensitive and/or personal data [86]. This will require more dictionaries and much, much larger data sets. Finally, while we used AVA to remove ambiguous terms and thereby pinpoint implicit gendered language, we do not necessarily advocate for use of AVA as a removal tool. AVA can be used to highlight or even heighten ambiguity, i.e., by adding AVA terms or swapping unambiguous synonyms with AVA terms, depending on the goal.

## 8.4 LIMITATIONS AND FUTURE WORK

Many of our analyses were limited by our chosen methods and materials. We have already discussed the issues with the dictionaries. The automated content analysis method is not without limitations, too, as it may miss instances of implicit bias, even with appropriate dictionaries and settings. We attempted to account for this with our follow-up analyses, i.e., our manual contextual analysis for Study 1, which led to the creation of AVA. Yet, we acknowledge that instances of implicit bias could have been missed. More work on implicit bias, ideally triangulating our strategies and findings, will be needed. The data sets are relatively small in terms of word count. This prevented us from carrying out other ML analyses, such as implicit bias via word embeddings, for which truly "big" data sets of millions or billions of words, if not utterances, would be needed. The data sets also do not represent diverse interactions with VAs, which, as we have discussed, likely affected our analyses. The AVA term list is relatively small and tied to the dictionaries and data sets used in this study; future work will need to expand on it, such as in focus groups with technologists or using crowdsourcing, both that include gender-diverse participant groups. We also did not use n-gram analyses, i.e., using a sequence of terms to predict the likelihood of the next in the sequence being biased, which can be done in future work.

## 9   Conclusion

We live in a gendered world that is expressed through and embedded, even in subtle ways, within language. AI-powered agents, interfaces, and spaces that we can communicate with through natural, verbal forms of language are on the rise. In tandem, an array of actors, some with great reach and power, are creating and providing materials to train the natural language abilities of these systems. This is a social good, with caveats. As this work shows, there are implicit gender biases in these data sets. Detection is the first step; next, we need to map out what forms of gendered language should be represented, with rigour and community engagement. Together, we can transcend the masculinities status quo.

**ACKNOWLEDGMENTS**

This work was funded by an Engineering Academy Young Scientist Encouragement Award (Tokyo Institute of Technology).